\documentclass{article}
\PassOptionsToPackage{numbers, compress}{natbib}
\usepackage[preprint]{neurips_2024}

\usepackage[utf8]{inputenc} 
\usepackage[T1]{fontenc}    
\usepackage{hyperref}       
\usepackage{url}            
\usepackage{booktabs}       
\usepackage{amsfonts}       
\usepackage{nicefrac}       
\usepackage{microtype}      
\usepackage{xcolor}         
\usepackage{graphicx}
\usepackage{wrapfig}
\usepackage{algorithm}
\usepackage{algpseudocode}
\usepackage{afterpage}
\usepackage{amsmath} 
\usepackage{caption}
\usepackage{enumitem}
\usepackage{orcidlink}
 
\title{LIPE: Learning Personalized Identity Prior for Non-rigid Image Editing}

\author{%
  Aoyang Liu\textnormal{\textsuperscript{1}}
  \quad Qingnan Fan\textnormal{\textsuperscript{2}} 
  \quad Shuai Qin\textnormal{\textsuperscript{2}}
  \quad Hong Gu\textnormal{\textsuperscript{2}} 
  \quad Yansong Tang\textnormal{\textsuperscript{1}$^{\dagger}$} \\
 \quad 
 \textnormal{\textsuperscript{1}Tsinghua University \,\,\textsuperscript{2}VIVO } \\
 \quad
 \texttt{lay22@mails.tsinghua.edu.cn}, \texttt{fqnchina@gmail.com}, \\ 
 \quad
 \texttt{tang.yansong@sz.tsinghua.edu.cn}
 }

\begin{document}

\maketitle

\begin{center}
  \centering
  \includegraphics[width=\textwidth]{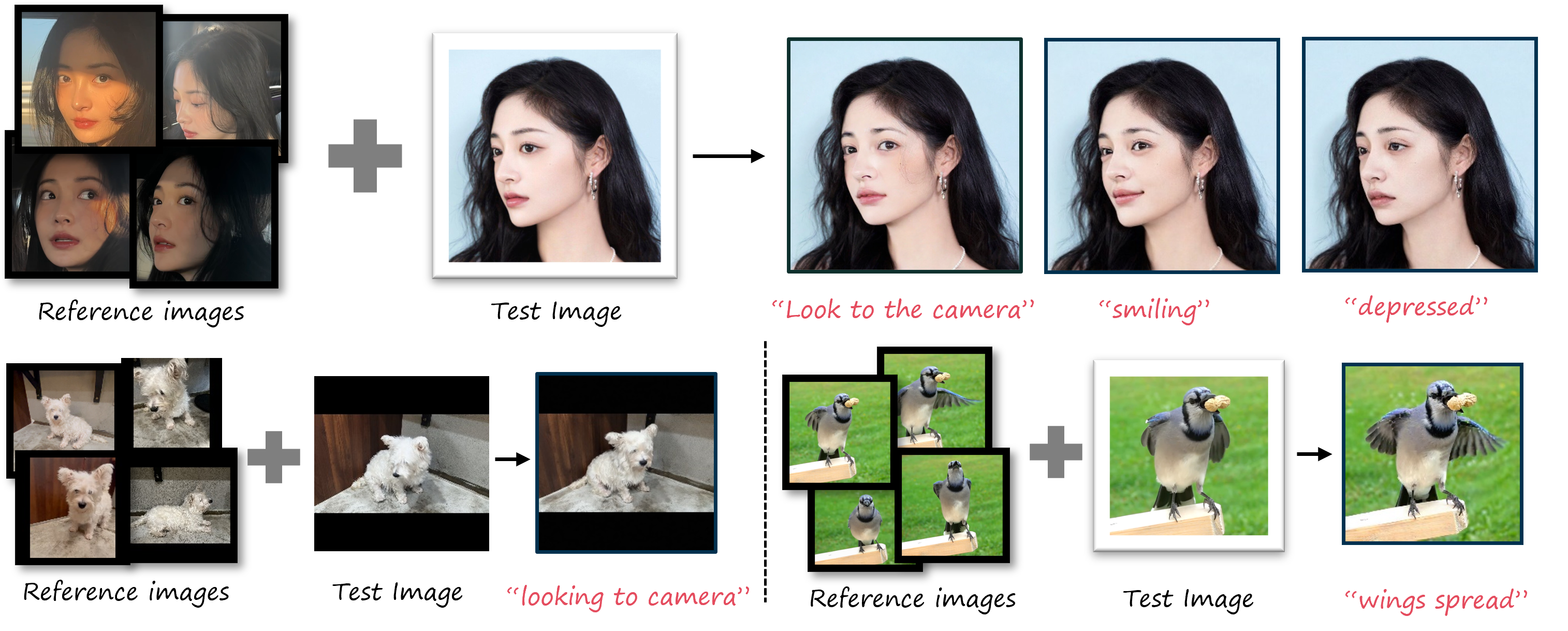}
  \captionof{figure}{Given a few reference images of the same identity, our framework learns a personalized identity prior and applies diverse non-rigid image editing for a test image guided by a textual description, leading to high identity-preserved edited results. 
  }
  \label{fig:teaser}
\end{center}

\let\thefootnote\relax\footnotetext{\scriptsize{$^{\dagger}$Corresponding author}}

\begin{abstract}
Although recent years have witnessed significant advancements in image editing thanks to the remarkable progress of text-to-image diffusion models, the problem of non-rigid image editing still presents its complexities and challenges. Existing methods often fail to achieve consistent results due to the absence of unique identity characteristics. Thus, learning a personalized identity prior might help with consistency in the edited results. In this paper, we explore a novel task: learning the personalized identity prior for text-based non-rigid image editing. To address the problems in jointly learning prior and editing the image, we present LIPE, a two-stage framework designed to customize the generative model utilizing a limited set of images of the same subject, and subsequently employ the model with learned prior for non-rigid image editing. Experimental results demonstrate the advantages of our approach in various editing scenarios over past related leading methods in qualitative and quantitative ways.
\end{abstract}

\section{Introduction}
\label{sec:intro}

Non-rigid image editing, such as altering the subject's posture, expression, or view angles, is a longstanding problem in computer vision and graphics \cite{oh2001image,kawar2023imagic}. It meets a wide spectrum of requirements in our daily life: adjusting the subject (e.g., human, pets, everyday object) in the captured image before sharing it on social media platforms \cite{pan2023drag}.

However, this is challenging due to the necessity of generating an outcome consistent with the original subject \cite{cao2023masactrl}.
Given the redundancy of images for the same identity from our smartphone's album and Internet, leveraging these photos could be highly beneficial in making an identity-preserved result in non-rigid image editing.
Therefore, in this work, we target a novel problem setting: 
\textit{given a small set (3-5) of reference images from the same identity, can we learn a personalized identity prior that facilitates the non-rigid editing of a test image while maintaining the unique properties of the identity?}

This problem has not been fully explored, and the closest works to ours are \cite{cao2023masactrl,kawar2023imagic,hertz2022prompt,parmar2023zero}, which adopt a general domain prior from the large-scale text-to-image (T2I) model for one image editing. These approaches underperform in our task for mainly two reasons.
1) Despite impressive results powered by the remarkable generative capabilities of the T2I model, they usually fail to faithfully preserve the identity's characteristics in the input image due to the lack of personalized identity prior.
2) They mainly rely on the less controllable text prompts with inversion \cite{song2020denoising,mokady2023null,huberman2023edit} for image editing, hence tend to modify the unwanted image regions. In addition, some recent works \cite{nitzan2022mystyle,zeng2023mystyle++} customize the personalized face prior with at least dozens of face images for portrait image editing on pretrained StyleGAN \cite{gal2022stylegan}, requiring more reference images ($\sim$100) and incapable of achieving non-rigid editing of more general subjects beyond faces.

A viable solution to this task involves integrating the personalizing generative prior method with the non-rigid image editing technique. However, previous personalizing generative priors methods excel in creating customized subjects but falter in generating subjects' fine-grained actions that align with text prompts, which are not suitable for image editing task. Additionally, when modifying the target subject, directly using the generative model to edit images often alters other attributes—such as background, lighting, and appearance—which are intended to remain constant.

To address the aforementioned issues, this study introduces a novel, unified solution based on a pre-trained T2I model termed \textbf{LIPE}, standing for \textbf{L}earning personalized \textbf{I}dentity \textbf{P}rior for non-rigid image \textbf{E}diting. For personalized identity prior learning, we identify that the critical element is utilizing the high-quality detailed text-image pairs to augment the model's capacity for depicting fine-grained actions or views. To achieve this, we devise a pipeline that leverages the large language-and-vision assistant and large language model to augment existing data.  For precisely controlling edited images, we find that introducing some masks referring to the target is beneficial. However, acquiring a mask covering the target subject before generating the final image is challenging. To overcome this, we utilize the property that attention maps can outline the general areas of the objects in the generated image in the diffusion model. Based on this, we propose an editing paradigm called \textit{Identity-aware mask blend (NIMA)}. NIMA automatically extracts an identity-aware mask and uses it to blend the latents of the target and source images during the denoising process, thereby allowing for more controlled non-rigid image editing.

To evaluate the effectiveness of our framework and facilitate further research, we introduce a new dataset specifically tailored to this novel task. This dataset encompasses a diverse range of classes, spanning from everyday objects (cups and toys), animals (cats and dogs) to human faces. For each class, there is one image for testing edits and several images for learning identity prior. This comprehensive dataset enables a thorough assessment of model quality. We conduct experimental comparisons between our proposed approach and related baselines. Experimental results demonstrate that our approach excels in preserving the original identity while achieving the desired edit effect.

We summarize our key contributions as follows:  (1) We introduce a \textbf{novel task}:
personalized identity prior for text-based non-rigid image editing; (2) We propose a \textbf{novel solution}, namely LIPE, short for \textbf{L}earning personalized \textbf{I}dentity prior and \textbf{P}racticing non-rigid image \textbf{E}diting, to tackle the associated technical challenges effectively; (3) We establish a \textbf{new dataset} to facilitate a comprehensive assessment of related works, aiming to advance research progress within the community.
\section{Method}
Due to space limitations, please refer to the appendix for the related work section. Given a few reference images of the same identity, our purpose is to learn a personalized identity prior and apply it to 
edit the subject's non-rigid properties (\textit{e.g.}, pose, viewpoint, expression) in the test image, guided by the text prompt while preserving the subject's characteristics with high fidelity and consistency. 

Our approach can be divided into two parts. First, we fine-tune the T2I model, SDXL \cite{podell2023sdxl} in implementation, to acquire the personalized identity prior using limited reference images like dreambooth \cite{ruiz2023dreambooth} (Subsection \ref{sec:edop}). After that, the non-rigid image editing via identity-aware mask blend is executed on the test image (Subsection \ref{sec:nima}). 

\subsection{Personalized Identity Prior}
\label{sec:edop}

\begin{figure}[tb]
  \centering
  \includegraphics[width=\textwidth]{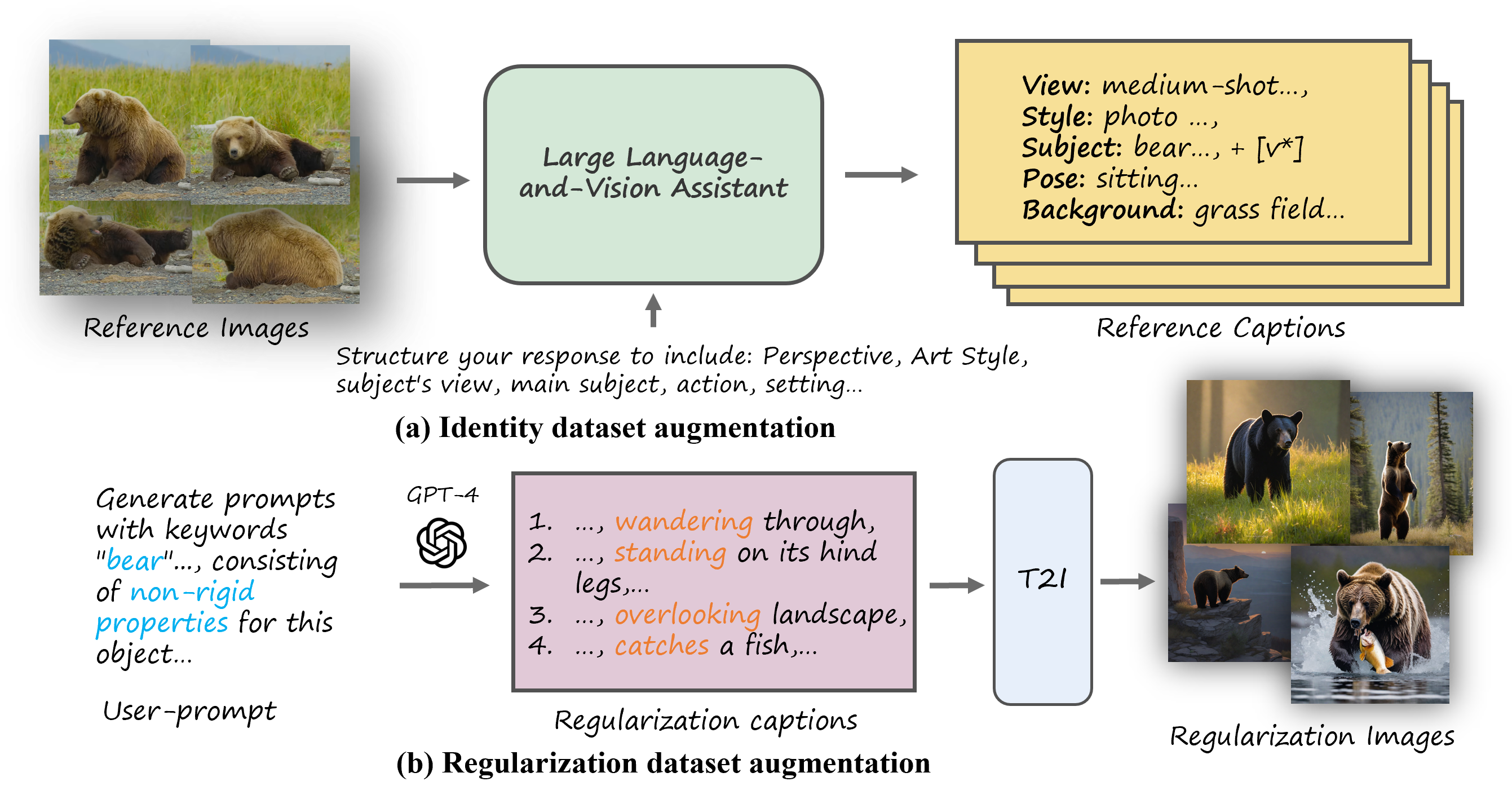}
  \caption{The pipeline for data augmentation in learning personalized identity prior. 
  (a) We make detailed editing-oriented captions for reference images by harnessing the large language and vision assistant. (b) We leverage the GPT-4 and pre-trained T2I model to generate diverse editing-oriented text-image pairs for the subject's class, which serves as the regularization dataset.}
  \label{fig:edop}
\end{figure}

In order to learn the personalized identity prior, many approaches \cite{gal2022image,ruiz2023dreambooth} proposed to fine-tune the pre-trained text-to-image (T2I) model $G$, provided a limited number ($N$) of reference images. However, they relied on less detailed text-image pairs during fine-tuning, leading to a personalized identity prior suitable for subject-driven generation but less effective for the non-rigid image editing task. Although the fine-tuned model could generate images of the target subject, it struggles to generate the same subject engaged in a given specific action. This indicates that the capability of fine-tuned models to depict specific non-rigid properties needs improvement. Therefore, we aim to enhance the model's capacity to comprehend the non-rigid properties of this identity and thus augment the model's efficacy in the downstream non-rigid image editing task. Based on this, we have adopted a new pipeline for data augmentation, enhancing the quality of data from the ID dataset to the regularized dataset, as detailed below and shown in Fig. \ref{fig:edop}.

\textbf{Editing-oriented data augmentation. }
Inspired by the recent progress \cite{betker2023improving} that fine-grained image captions are the key to generative models' controllability, to enable more accurate text-to-image mapping ability for non-rigid image properties, we propose to construct more delicate editing-oriented text-image pairs to learn the personalized identity prior. Our paired data consists of two main datasets: the identity and regularization datasets.

The identity dataset is derived from the reference images and designed to learn the unique characteristics of the target identity. Motivated by the large multi-modal models \cite{liu2024visual,ye2023mplug,chen2023sharegpt4v}, which is able to generate more flexible and customized captions than the traditional methods \cite{li2022blip,li2023blip,hu2022scaling}, we utilize the recent large language-and-vision assistant (LLaVa) \cite{liu2024visual} to generate the caption counterparts for the reference images. We have specifically designed a prompt for LLaVa in the hope that it can offer the caption we need, which includes accurate and comprehensive descriptions of the non-rigid properties of the subject, as well as other features of the image, illustrated in Fig. \ref{fig:edop}(a).

The regularization dataset is designed to address the image drift issue \cite{ruiz2023dreambooth}, where a diffusion model finetuned for a specific task tends to lose the general semantic knowledge and lack diversity. Different from the previous methods that solely relied on coarse prompts describing what the instance is, we produce finely detailed prompts delineating the non-rigid properties of the instance with the help of GPT-4 \cite {achiam2023gpt} and further utilize the recent large-scale T2I model, SDXL \cite{podell2023sdxl}, to generate the corresponding images, as shown in Fig. \ref{fig:edop}(b). The resultant identity and regularization datasets both serve as the database to learn the identity prior. Please see the supplementary material for the details of prompt design for the large language model.

\begin{figure}[tb]
  \centering
  \includegraphics[width=\textwidth]{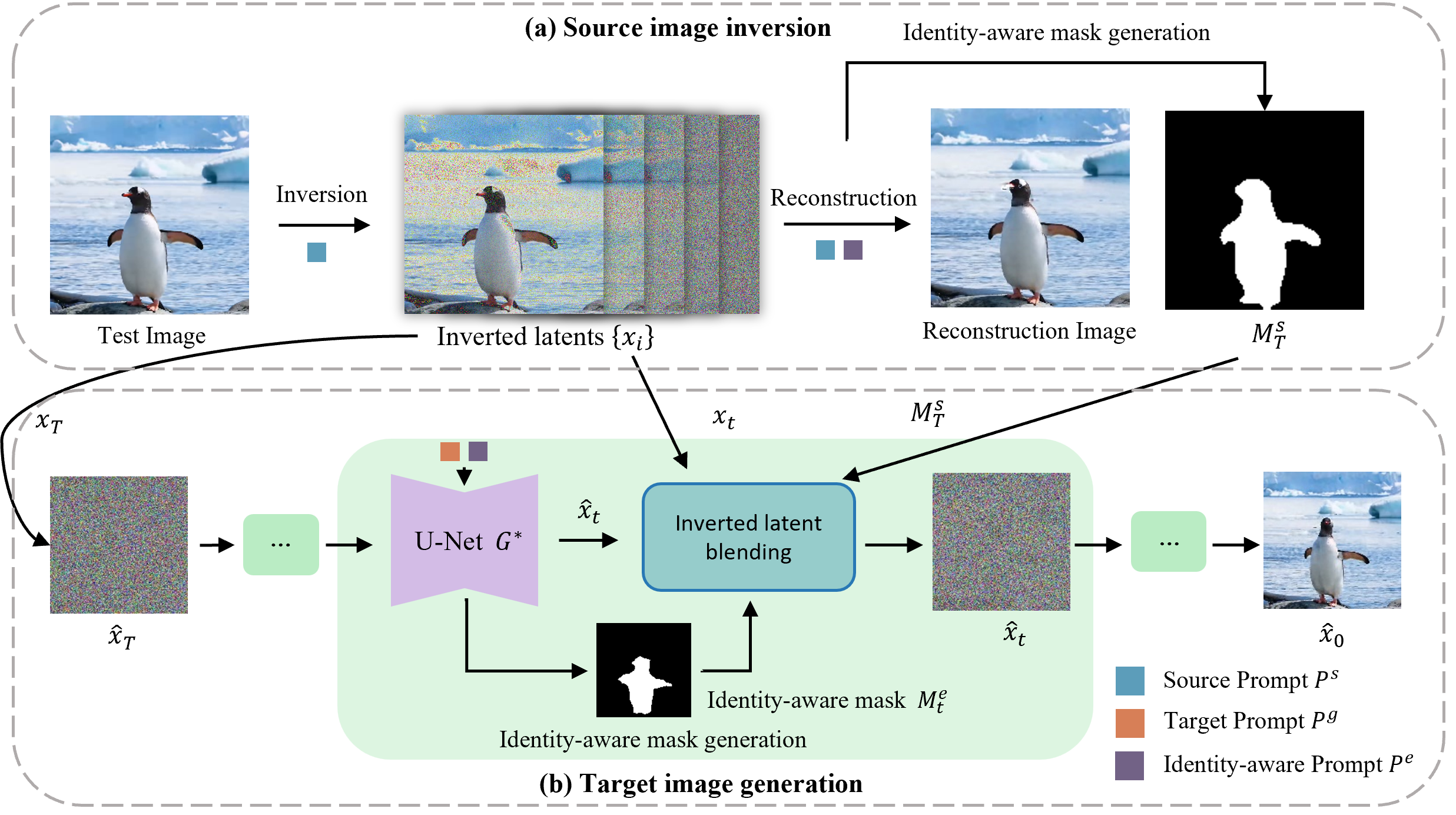}
  \caption{Illustration of Non-rigid Image editing via identity-aware MAsk blend (NIMA). (a) Given a test image, we first invert it to obtain the inverted latents $\{x_i\}$ for image reconstruction, to further obtain the subject mask $M^s$ for the source image. 
  (b) Afterward, to achieve non-rigid image editing, we generate the target image by blending the source $x_t$ and target $\hat{x}_T$ information with the generated masks ($M^s$, $M_t^e$).
  }
  \label{fig:nima}
\end{figure}

\subsection{Non-rigid Image Editing via Identity-aware Mask Blend}
\label{sec:nima}

Having learned the personalized identity prior, our objective shifts towards leveraging it for precise non-rigid image editing. To precisely control edited images, we leverage masks that represent the target objects before denoising the final outputs. To this end, we propose a non-rigid image editing pipeline that presciently extracts identity-aware masks before denoising the final output and blends the source and target information in the latent level for editing. 
The entire process is depicted in Fig \ref{fig:nima}. Initially, we invert the test image to acquire the inverted latents with the information of source prompt $P^s$ (which is obtained by a captioner in Subsection \ref{sec:edop}). Subsequently, we produce the target image from the inverted latent $\hat{x}_T$ using the target prompt $P^g$ and utilize masks to guide the editing. 

\begin{wrapfigure}{r}{0.5\textwidth}
  \includegraphics[width=0.5\textwidth]{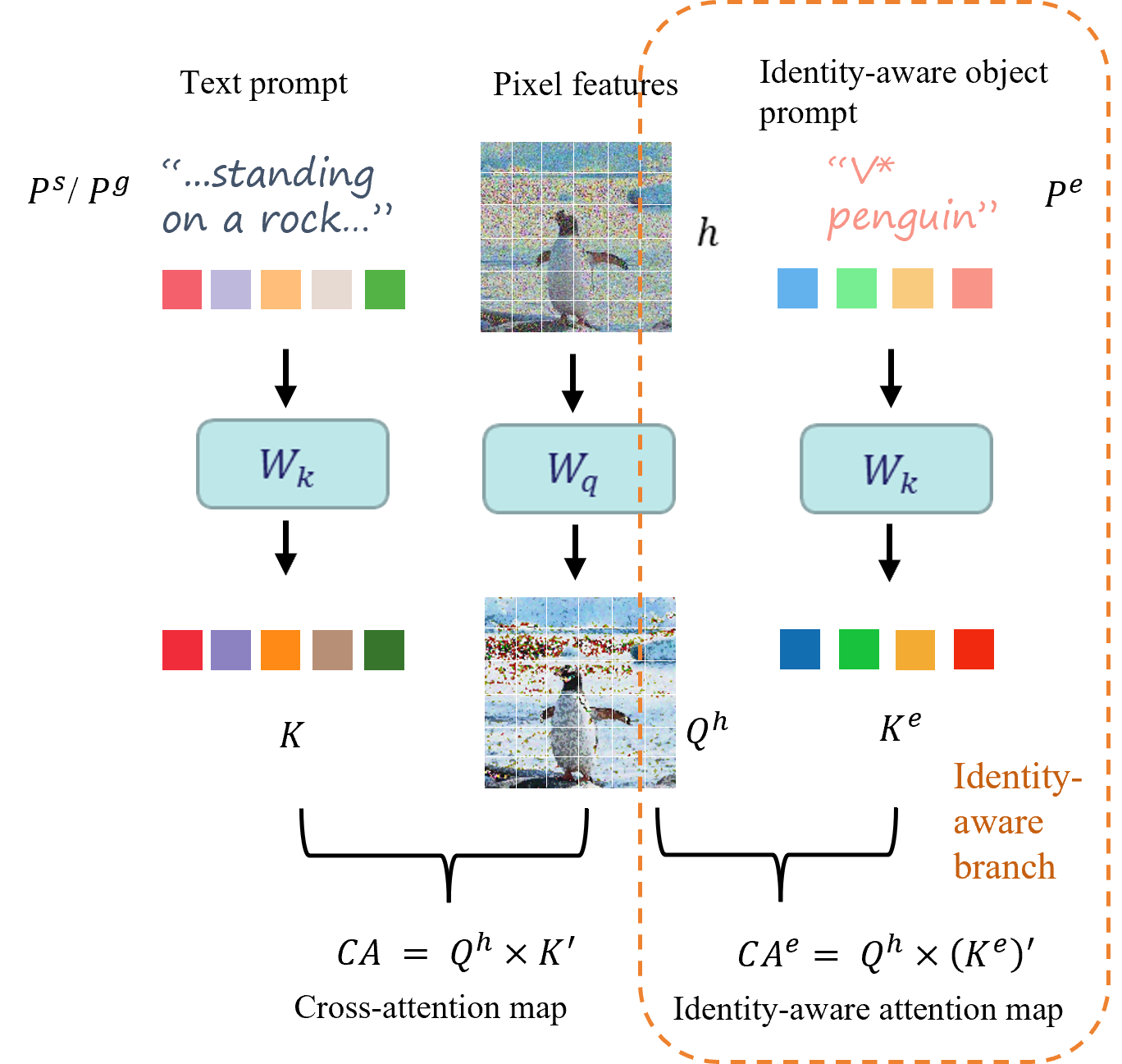}
  \caption{Identity-aware attention map.}
  \label{fig:mask1}
\end{wrapfigure}

\textbf{Identity-aware mask extraction.}
We aim to determine a precise mask for the target subject before generating the final target image. Inspired by the observation that cross-attention maps delineate areas corresponding to each semantic text token \cite{hertz2022prompt,tang2022daam}, we consider the cross-attention maps for some text tokens related to identity (referred to as identity-aware object prompt $P_e$ ) as an approximate mask. To implement this, we add an extra identity-aware branch in the diffusion denoising process to calculate the identity-aware attention maps, as illustrated in Fig \ref{fig:mask1}. To represent the identity, the identity-aware object prompt $P^e$ is often defined as a specific object, such as ``penguin'', ``bird'', or ``cat'' with the rare token $V^*$. Thus in the cross-attention module of the model, apart from computing the attention map between the intended text prompt ($P^s,P^g$) and the image, the attention map between the identity-aware text prompt $P^e$ and the image is also computed to obtain a mask. It is worth noting that the identity-aware branch does not interfere with the original generation process.

To be specific, we first average the identity-aware attention maps across all layers and all previous steps to obtain a rough mask $M^1$, formulated as:

\begin{equation}
    M^1_t = \frac{1}{(T-t) \times L}\sum_{\tau=T}^{t}\sum_{l=1}^{L}\text{CA}^{e}_{\tau,l},
\end{equation}
where $\text{CA}^{e}_{\tau,l}$ is the identity-aware attention map in layer $l$ and timestep $\tau$.

\setlength{\intextsep}{7pt}

Considering that self-attention maps reflect the layout and grouping regions \cite{hertz2022prompt,tumanyan2023plug,patashnik2023localizing,nguyen2024dataset,wang2023diffusion}, to refine the mask $M^1_t$, we multiply it with averaged self-attention maps $\overline{\text{SA}}$ to acquire the final identity-aware mask $M^2_t$. 
\begin{equation}
    M^2_t = \overline{\text{SA}}^{\lambda} \times M^1_t,
\end{equation}
where $\overline{\text{SA}}$ is the average self-attention maps across all layers and all previous steps and $\lambda$ is the hyper-parameter controlling the extent of sharpening.

In NIMA, we obtain two types of identity-aware masks $M^2_t$: source mask $M^s_T$ for the original test image and target mask $M_t^e$ for the target image. The source mask is obtained at the last step in the reconstruction process and the target mask is obtained during the denoising process dynamically. Target $M_t^e$ is updated at each denoising timestep $t$.

\textbf{Inverted latent blending.} At each timestep $t$, we utilize the two masks and inverted latents to control the generation process. First, mask $M_t$ is computed and serves as a mask guidance.
\begin{equation}
    M_t = M^s_T \cup M_t^e,
\end{equation}
The union of two masks can prevent the phenomenon of overlapping two subjects in a single image.

Unlike previous editing methods \cite{avrahami2022blended,avrahami2023blended,patashnik2023localizing}, we blend the two latents generated by the inversion process instead, avoiding the inconsistent foregrounds and noticeable boundaries. For the area outside the mask, we maintain the test image's inverted latent to preserve the original image's background information. 
\begin{equation}
    \hat{x}_T = M_t\hat{x}_T + (1-M_t)x_t,
\end{equation}

\begin{algorithm}[t]
\caption{Non-rigid Image editing via identity-aware MAsk blend (NIMA)}
\begin{algorithmic}[0]
\raggedright
    \State \textbf{Input:} Test image $I^s$, source prompt $P^s$, target prompt $P^g$, identity-aware prompt $P^e$, fine-tuned model $G^*$.
    \State \textbf{Output:} Edited image $I^e$.
    \State $x_0$ = Encoder($I^s$,$G^*$)
    \State $\{x_i\}$ = Inversion($x_0$,$G^*$,$P^s$) 
    \State $M^s,{I^{s}}'$ = Sampling($x_T$,$G^*$,$P^s$,$P^e$)   \Comment{reconstruction Image ${I^{s}}'$ and source mask $M^s$}
    \State $\hat{x}_T=x_T$ \\
    \For{$t=T$, $T-1$, $T-2$,..., $0$}
      \State $\hat{x}_{t-1},M_t^e=$ Denoise($\hat{x}_T,G^*$,$P^g$,$P^e$) \Comment{diffusion denoising step with extra branch}
      \State $M_t = M^s_T \cup M_t^e$
      \State $\hat{x}_{t-1} = M_t \hat{x}_{t-1} + (1-M_t) x_{t-1}$ \Comment{blend inverted latents}
    \EndFor
    \State $I^e$ = Decoder($\hat{x}_0$,$G^*$)
\end{algorithmic}
\end{algorithm}
\section{Experiments}
In this section, we first introduce the new dataset (LIPE) which is brought to assess different related methods (Subsection \ref{sec:dataset}). Then, we display the comprehensive comparisons with previous related methods in both qualitative and quantitative ways (Subsection \ref{sec:res}). Finally, we further analyze the role of each technical component from LIPE (Subsection \ref{sec:aba}). Please see the supplementary material for full implementation details and more experimental results.

\subsection{Dataset}
\label{sec:dataset}

We present a novel dataset, LIPE, designed specifically for this task.  LIPE comprises a total of 28 subjects, encompassing a diverse range from everyday objects and animals to human faces, facilitating the exploration of various non-rigid properties editing (e.g., poses, viewpoints, and expressions).
Each instance within the LIPE dataset is mostly accompanied by 3-5 reference images for learning the identity prior, along with one image designated for testing edits. Please see the supplementary material for more details about this dataset. 

\subsection{Comparisons with Previous Works}
\label{sec:res}
We mainly make comparisons with several current leading related works specializing in non-rigid image editing tasks \cite{cao2023masactrl,nitzan2022mystyle,kawar2023imagic}. Moreover, we also design a baseline combining two leading approaches \cite{cao2023masactrl,ruiz2023dreambooth} from identity prior learning and non-rigid image editing, denoted as DreamCtrl. DDPM inversion \cite{huberman2023edit} is employed for all methods with inversion process.

We elaborate on all the related methods below:
\begin{itemize}[leftmargin=*]
\item[$\bullet$] \textbf{MasaCtrl} \cite{cao2023masactrl}. It performs the non-rigid image editing on the T2I model without any personalized identity prior. To better enhance the generation quality and ensure a fair comparison, source prompt $P_s$ is utilized in the inversion branch instead of null. This method justifies the necessity of adding personalized identity prior to the non-rigid image editing task. 
\item[$\bullet$] \textbf{Imagic} \cite{kawar2023imagic}. It optimizes the whole network and textual embedding to reconstruct the source image, and then interpolates the original embedding and optimized embedding to acquire the non-rigid edits. Similar to MasaCtrl \cite{cao2023masactrl}, Imagic \cite{kawar2023imagic} also only utilize the general prior for the non-rigid editing.
\item[$\bullet$] \textbf{DreamCtrl}. Because personalized identity prior for text-based non-rigid image editing is a novel task, we design a baseline merging two leading methods in identity prior learning and non-rigid image editing, DreamBooth \cite{ruiz2023dreambooth} and MasaCtrl \cite{cao2023masactrl}. Specifically, the T2I model is fine-tuned in the DreamBooth \cite{ruiz2023dreambooth}'s way, and then image editing is performed on the fine-tuned model in the MasaCtrl \cite{cao2023masactrl}'s way. DreamCtrl shares the same regularized dataset with LIPE to enhance this method's learning but has no editing-oriented captions.
\item[$\bullet$] \textbf{MyStyle} \cite{nitzan2022mystyle}. It learns a personalized identity prior from about 100 reference images and leverages it for human face image editing based on a pre-trained StyleGAN \cite{karras2019style}. It relies on several pre-extracted semantic directions from typical StyleGAN \cite{karras2019style} to modify the latent to achieve semantic editing. During comparison, we use directions that align closely with our target prompt's meaning. To improve the reconstruction of the test image, we also follow the optional fine-tuning process on the test image. Moreover, because this method only focuses on the human face domain, all evaluations are considered only on human face results. 
\end{itemize}

\begin{figure}[!tb]
  \centering
  \includegraphics[width=\textwidth]{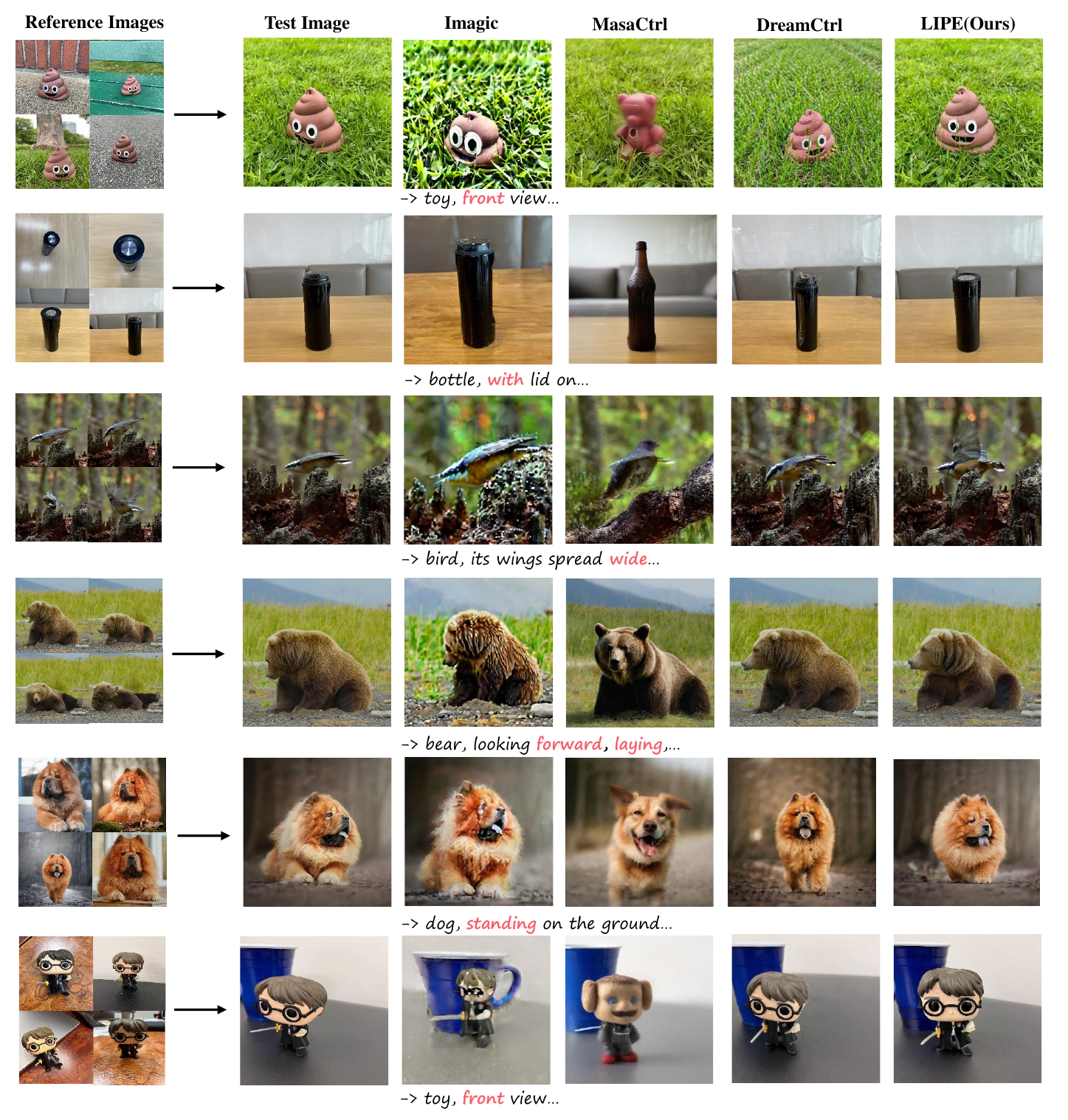}
  \caption{\textit{Comparisons with previous work on general objects.} The red font highlights the editing directions. Left to right: Reference images, Test image, Imagic \cite{kawar2023imagic}, MasaCtrl \cite{cao2023masactrl}, DreamCtrl, and Our method.}
  \label{fig:res1}
\end{figure}

\subsubsection{Qualitative Results on General Objects} 
The results on general objects are shown in Fig \ref{fig:res1}. Our method exhibits the flexible capacity for non-rigid transformation and high-quality preservation of background information, outperforming other approaches. Based on these outcomes, we claim that the personalized identity prior is essential for preserving the identity's characteristics in non-rigid image editing, as the edited results of Imagic \cite{kawar2023imagic} and MasaCtrl \cite{cao2023masactrl} indicate. Compared to merely amalgamating the two existing methodologies—specifically, DreamCtrl—our approach significantly enhances the generation of specified actions while preserving the integrity of the background.

\begin{figure}[tb]
  \centering
  \includegraphics[width=\textwidth]{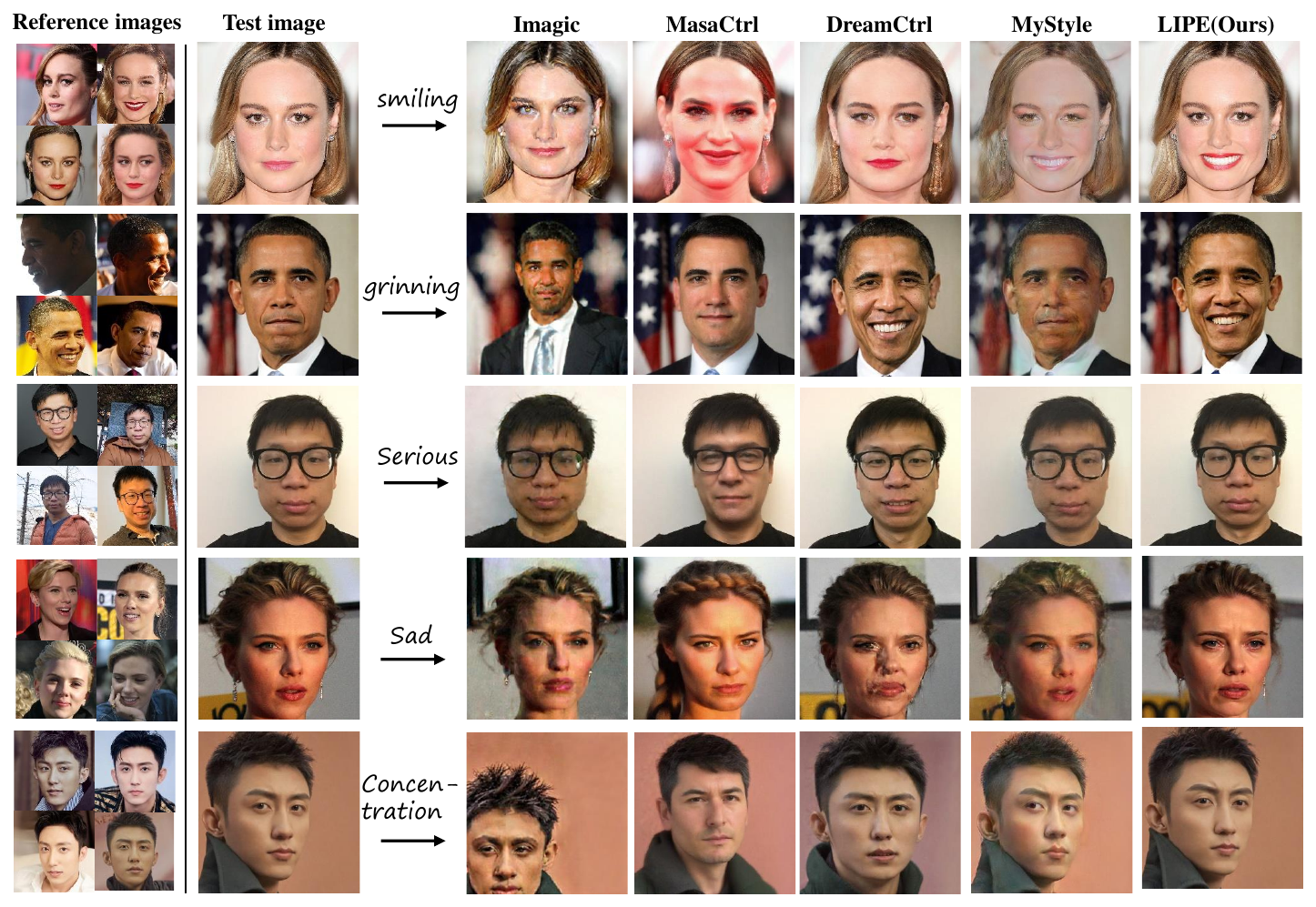}
  \caption{\textit{Comparisons with previous work on human face.} Left to right: Reference images, Test image, Imagic \cite{kawar2023imagic}, MasaCtrl \cite{cao2023masactrl}, DreamCtrl, MyStyle \cite{nitzan2022mystyle}, and Our method. The text on the arrow represents the target prompt utilized during the editing process, aligning with the semantic direction of the StyleGAN\cite{gal2022stylegan} utilized in MyStyle\cite{nitzan2022mystyle}.}
  \label{fig:res2}
\end{figure}

\subsubsection{Qualitative Results on Human Faces} 
The results on human faces are shown in Fig \ref{fig:res2}. Editing human faces poses greater challenges compared to general subjects due to the complex and subtle nature of their unique characteristics. And the experimental results on human faces further corroborate the claim from our study on general subjects that only general prior is not enough for non-rigid editing. Moreover, our approach demonstrates better background preservation and efficiently outputs prompt-compliant images compared to DreamCtrl. MyStyle \cite{nitzan2022mystyle} struggles to perform effectively when trained on very limited data (less than 5 images, see 2,3,4 rows) and to preserve the information of the original image when trained on few images (less than 10 images, see 1,5 row). 

\setlength{\tabcolsep}{10pt}
\begin{table}[t]
  \caption{User Study. The edited results are evaluated from three detailed aspects: Identity Consistency, Background Consistency, and Editing Satisfaction. And the overall score is the average of the three previous scores. MyStyle \cite{nitzan2022mystyle} is only evaluated on human faces.}
  \label{tab:user}
  \centering
  \begin{tabular}{ccccc}
    \toprule
     & Identity $\uparrow$  & Background $\uparrow$   & Editing  $\uparrow$  & Overall $\uparrow$   \\
    \midrule
    Imagic \cite{kawar2023imagic} & 0.4545 & 0.5908 & 0.1818 & 0.4090 \\
    MasaCtrl \cite{cao2023masactrl} &0.1785 & 0.4714  & 0.3964 & 0.3488 \\
    DreamCtrl & 0.7535 & 0.6892  & 0.6464 & 0.6964 \\
    MyStyle \cite{nitzan2022mystyle} & 0.7500 &  0.8100  & 0.4400 & 0.6666 \\
    \midrule
    \textbf{LIPE(Ours)} & \textbf{0.9464} & \textbf{0.8857} & \textbf{0.7357} & \textbf{0.8559} \\
    \bottomrule
  \end{tabular}
\end{table}

\setlength{\tabcolsep}{10pt}
\begin{table}[t]
  \caption{Quantitative Evaluation. We measure the quality of edited results in 4 metrics utilizing the DINO \cite{oquab2023dinov2} and CLIP \cite{radford2021learning}. MyStyle \cite{nitzan2022mystyle} is only evaluated on human faces.}
  \label{tab:quan}
  \centering
  \begin{tabular}{ccccc}
    \toprule
     & DINOv2-I$\uparrow$ & CLIP-I$\uparrow$  & CLIP-T$\uparrow$ & CLIP-D$\uparrow$   \\
    \midrule
    Imagic \cite{kawar2023imagic} & 0.6961 & 0.7356 & 0.2310 & 0.0138 \\ 
    MasaCtrl \cite{cao2023masactrl} & 0.6027 &  0.7425  & \textbf{0.2628} & 0.0178\\
    DreamCtrl & 0.7804 &   0.8517 & 0.2304 & 0.0433\\
    MyStyle \cite{nitzan2022mystyle} &  0.8148  & 0.7371  &  0.1892 & 0.0251 \\
    \midrule
    \textbf{LIPE(Ours)} & \textbf{0.8848} & \textbf{0.9247} & 0.2331 & \textbf{0.0807} \\
    \bottomrule
  \end{tabular}
\end{table}

\subsubsection{User study} 
To evaluate the effectiveness of algorithms from the perspective of human aesthetics, we conducted a user study on the LIPE dataset. We engaged 30 users to evaluate the outcomes based on three key aspects: (1) Identity Consistency: how effectively the edited result retains the distinctive characteristics of the original identity. (2) Background Consistency: how completely the edited result maintains the background of the test image. (3) Editing Satisfaction: whether the required non-rigid transformation is successfully achieved. Users were asked to assign scores from choices in [0,0.5,1], where higher scores indicate better performance. Subsequently, the scores for each aspect were averaged across users for each method. The overall aspect score was then calculated as the mean of the scores for the three aspects. The results are reported in Table \ref{tab:user}.
Our method demonstrated significant advantages over other methods across all three aspects. 

\subsubsection{Quantitative Evaluation}
We quantitatively evaluated our method with other methods, employing four metrics to assess identity preservation and text-image alignment. The DINOv2-I metric calculates the pairwise cosine similarity of test images and edited images in the space of DINOv2 \cite{oquab2023dinov2} on average. Similarly, the CLIP-I metric computes the pairwise cosine similarity of two images in CLIP \cite{radford2021learning} space. These two metrics serve as indicators of identity preservation. CLIP-T, introduced by \cite{ruiz2023dreambooth}, evaluates text-image alignment by computing the pairwise cosine similarity between the edited images and the target prompts in the CLIP \cite{radford2021learning} space, which often measures the model's performance for generation task. The CLIP text-image direction similarity metric (CLIP-D), introduced by \cite{gal2022stylegan}, assesses the alignment of text changes with corresponding image alterations. It often serves as a metric for image editing tasks \cite{nguyen2024visual}. Results in Tab. \ref{tab:quan} indicate LIPE achieves superior performance in identity preservation and exhibits strong alignment between text descriptions and image changes. MasaCtrl \cite{cao2023masactrl} displays higher CLIP-I scores, likely due to its tendency to \textit{generate} the image aligning with the target prompt instead of \textit{editing} the image.

\subsection{Ablation Study}
\label{sec:aba}

\begin{figure}[tb]
  \centering
  \includegraphics[width=\textwidth]{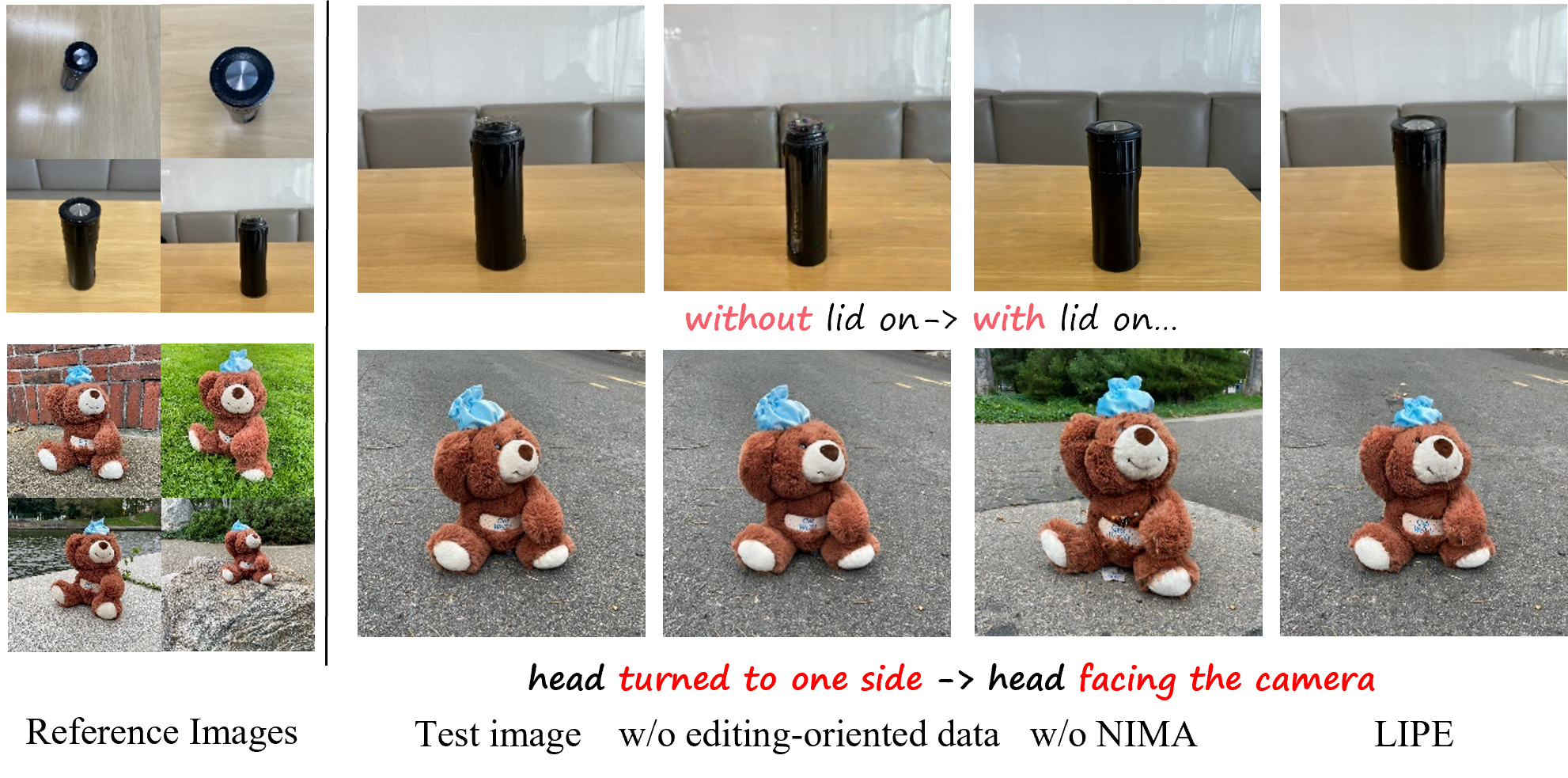}
  \caption{Ablation Study. In the ablation study, we confirmed the effectiveness of the two strategies in our method. Adding action-descriptive captions while learning the personalized identity prior helps the model comprehend non-rigid properties to assist with editing. Obtaining target mask information in advance during the editing process helps the method control the outcomes more precisely.}
  \label{fig:aba}
\end{figure}

We conducted an ablation study to delineate the roles of detailed captions and masks in our method for non-rigid image editing. As shown in the Fig. \ref{fig:aba}, without detailed action-descriptive captions to train the model's priors, it is challenging to make the subject perform specific actions. Moreover, the mask precisely helps control the consistency of the background before and after editing.

\section{Conclusion}
We have introduced a novel problem: learning a personalized identity prior and utilizing it for non-rigid image editing. To address this challenge, our method (LIPE) leverages a set of reference images to learn a personalized identity prior, which greatly benefits the editing task. Subsequently, non-rigid image editing is performed using identity-aware mask blending to achieve precise editing. Experiments show that our method outperforms previous methods in both qualitative and quantitative ways. We believe that our work opens up new possibilities in the realm of non-rigid image editing and serves as a catalyst for future research in this area.

\bibliographystyle{acl_natbib}
\bibliography{neurips_2024}
\medskip

\newpage
{
\small
\appendix
\section{Appendix / supplemental material}
We start with section \ref{sec:related_work} to introduce recent related research. Subsequently, section \ref{sec:background} shows background about the diffusion model and inversion technique. Next, in Section \ref{sec:method} and Section \ref{sec:exp}, we show the details about our method and experiments, respectively. And section \ref{sec:add results} provides additional examples, demonstrating our method across diverse subjects and backgrounds. Finally, section \ref{sec:limitation} discusses the limitations of our method.
\section{Related Work}

\label{sec:related_work}

\subsection{Text-to-image generation}
During the early stages, text-to-image models primarily relied on generative adversarial networks (GANs) with a generator-discriminator structure, trained on small-scale datasets \cite{goodfellow2014generative,reed2016generative,zhang2017stackgan,xu2018attngan,li2019controllable}. Subsequently, auto-regressive methods were explored on large-scale data \cite{ramesh2021zero,ding2021cogview,wu2022nuwa,yu2022scaling}. However, these methods faced challenges of high computation costs and sequential error accumulation \cite{zhang2023text}. Recently, diffusion models \cite{ho2020denoising,song2020denoising} have gradually become prominent in this field, leveraging large-scale models on large-scale data \cite{rombach2022high,nichol2021glide,ramesh2022hierarchical,saharia2022photorealistic,gu2022vector}. These models denoise images by modeling images from Gaussian distributions through a Markov-chain process and exhibit significant capabilities in generating photorealistic images. Presently, several works \cite{peebles2023scalable,gupta2023photorealistic,sora} employ the transformer architecture \cite{vaswani2017attention} as the backbone of the diffusion model instead of U-Net \cite{ronneberger2015u}, aiming to scale it and produce high-quality images or videos.

\subsection{Text-guided image editing}
Before diffusion models emerged as the predominant approach in image generation tasks, numerous methods were devised to manipulate images within the latent space of Generative Adversarial Netwofrks (GANs) \cite{alaluf2022hyperstyle,patashnik2021styleclip,shen2020interpreting,shen2021closed,nam2018text,li2020manigan,patashnik2021styleclip,zhang2023sine}. They utilize the decoupling properties of features in the StyleGAN's latent space to discover some editing directions. Apart from GAN-based methods, certain techniques employ other deep learning-based systems for image editing \cite{bar2022text2live,chang2022maskgit}.

More recently, with the popularity of diffusion models, various approaches have been devised to utilize diffusion models for image editing \cite{brooks2023instructpix2pix,kim2022diffusionclip,nichol2021glide,geng2023instructdiffusion,meng2021sdedit,avrahami2022blended,couairon2022diffedit,hertz2022prompt,tumanyan2023plug,cao2023masactrl,parmar2023zero}. Some methods focus on general type of edits \cite{hertz2022prompt,brooks2023instructpix2pix,couairon2022diffedit}, while others specialize in image style transfer \cite{tumanyan2023plug,zhang2023sine}, inpainting \cite{avrahami2022blended,xie2023dreaminpainter}, or object replacement \cite{parmar2023zero}.  Presently, the non-rigid edits become a noteworthy direction \cite{cao2023masactrl,kawar2023imagic}. Imagic \cite{kawar2023imagic} optimizes the text embedding and model's weight to hold the original image's information. Masactrl \cite{cao2023masactrl} copies the key and value features from the original generation process to keep the color and texture information of the original subject in a training-free way. However, it is hard for these approaches to achieve non-rigid edits while holding the original subject since they only use the general prior without specific identity information. Therefore, in this work, we consider adding the personalized identity prior to preserve the key features of the target subject from the test image.

\subsection{Personalized Generative Prior}
The problem of learning a personalized prior with the pre-trained generative model was first researched on GANs \cite{alaluf2022hyperstyle,roich2022pivotal,pan2021exploiting,bau2020semantic}. Based on pivotal tuning \cite{roich2022pivotal}, MyStyle \cite{nitzan2022mystyle} learns the domain prior trained on the large-scale face dataset (FFHQ \cite{karras2019style}) and the identity prior on approximately 100 face images for portrait image editing.

Different from MyStyle \cite{nitzan2022mystyle}, our work focuses on general subject editing and only requires 3-5 images to learn the personalized identity prior.

In recent years, many methods have tried to personalize the text-to-image diffusion models by fine-tuning text embedding \cite{gal2022image}, full weights\cite{ruiz2023dreambooth}, a subset of cross-attention layers \cite{kumari2023multi} or singular values \cite{han2023svdiff}. Other works incorporate extra adapters to receive a reference image and generate the image with the same subject \cite{ye2023ip,li2024blip,li2023photomaker,wang2024instantid}. Based on dreambooth \cite{ruiz2023dreambooth}, DreamEdit \cite{li2023dreamedit} first learns a personalized identity prior and then employs it to do editing. However, it only handles rigid image editing tasks, such as subject replacement and subject addition. To the best of our knowledge, ours is the first paradigm to apply the personalized identity prior to a non-rigid image editing task for general subjects, based on the state-of-the-art T2I model.

\section{Background}
\label{sec:background}
\subsection{Diffusion model}
\label{sec:pre}
Diffusion models \cite{dhariwal2021diffusion,ho2020denoising,rombach2022high,nichol2021improved,song2020denoising} are probabilistic generative models that learn data distributions by defining a Markov chain of diffusion steps and reversing them. During the diffusion forward process, Gaussian noise $\epsilon$ is added to an image sample $x_0$, resulting in a noisy sample $x_t$ at time step $t$, mathematically expressed as:
\begin{equation}
    q(x_t|x_{0}) = \mathcal{N}(x_t;\sqrt{\bar{\alpha_t}}x_{0},(1-\bar{\alpha_t}) I)
\end{equation}
where $\alpha_t$ is the predefined noise adding schedule and $\bar{\alpha_t}=\prod_{i=1}^t \alpha_i$. The diffusion models then reverse this process by denoising $x_{t-1}$ from the next step $x_t$ iteratively to recover the original image sample $x_0$. This denoising process is modelled as follows:

\begin{align}
    p\left({x}_{t-1} \mid {x}_t \right) &= \mathcal{N} ({x}_{t-1} ;\boldsymbol{\mu}_t ({x}_t), \sigma_t)
    \\ 
    \boldsymbol{\mu}_t({x}_t) &= \frac{1}{\sqrt{\alpha_t}}(x_t - \frac{1-\alpha_t}{\sqrt{1-\bar{\alpha_t}}}\epsilon_\theta(x_t,t))
\end{align}
where mean $\boldsymbol{\mu}_t({x}_t)$ can be solved by neural network $\epsilon_\theta(x_t,t)$ and $\sigma_t$ is only determined by timestep $t$. The neural network is typically trained using squared error loss to predict the added noise:

\begin{equation}
    \label{eq:loss}
    \mathcal{L}_{\text(loss)} = E_{x_0,\epsilon,t}(||\epsilon - \epsilon_{\theta}(x_t,t)||_2^2)
\end{equation}

Recent years, diffusion models have advanced to condition on text \cite{kim2022diffusionclip,nichol2021glide,ramesh2022hierarchical}, and both the forward and denoising process are conducted in latent space \cite{rombach2022high}. 

\subsection{Inversion}
Inversion is a technique to obtain the initial noise that can generate the given image through the forward pass in diffusion models. It is an essential component for real-world image editing, which stems from the DDIM sampling \cite{song2020denoising}.

\begin{equation}
    \label{eq:ddim}
    x_{t-1} = \sqrt{\bar{\alpha_{t-1}}}(\frac{x_t-\sqrt{1-\bar{\alpha_t}}\epsilon_{\theta}(x_t,t)}{\sqrt{\bar{\alpha_t}}}) + \sqrt{1-\bar{\alpha_{t-1}}}\epsilon_{\theta}(x_t,t)
\end{equation}
This formulation can be seen as a neural ODE \cite{song2020denoising} and a determistic precedure. Therefore, within enough discretization steps, we can reverse the generation process, which is called DDIM Inversion or DDIM encoding.

\begin{equation}
    {x_{t+1}} = \sqrt{\bar{\alpha_{t+1}}} \cdot {x_0} +\sqrt{1-\bar{\alpha_{t+1}}}{\epsilon(x_t)}
\end{equation}

However, 
DDIM inversion does not reconstruct well the given image when deploying the class-free guidance \cite{ho2022classifier,mokady2023null} or modified text prompt \cite{huberman2023edit}, hence more inversion methods emerged for further improvements \cite{mokady2023null,huberman2023edit,ju2023direct,miyake2023negative}.
\section{Method details}

\label{sec:method}
\subsection{Prompt design in data augmentation}
\subsubsection{Prompt for LLaVA}
Our object is to make the captions include accurate and comprehensive descriptions of the non-rigid properties of the subject. In addition, incorporating several descriptive captions unrelated to the subject helps prevent the model from over-fitting to these irrelevant "backgrounds." The prompt we designed is shown in Tab. \ref{tab:prompt1}.

\begin{table}[tb]
  \caption{Prompt on LLaVA for detailed descriptions generation.}
  \label{tab:prompt1}
  \centering
  \begin{tabular}{p{\textwidth}}
    \toprule
    Enhance your image captioning skills by providing detailed captions. Structure your response to include: Perspective (e.g., headshot, medium-shot), Art Style (e.g., digital artwork, Ghibli-inspired screencap, photo), subject's view(facing to the camera), Main Subject (e.g., a person, an object, a woman), Action (if any, running), Attire Details, Hair Description, and Setting. Aim for concise yet descriptive captions, ideally under 70 words. For example: 'Close-up, Ghibli-style digital art, a young lady, gazing thoughtfully, adorned in a vintage dress, her hair styled in a loose bun, amidst a whimsical forest backdrop.' Now tell me the caption of this image. \\ 
    \bottomrule
  \end{tabular}
\end{table}

\subsubsection{Prompt for GPT-4}
We aim to leverage GPT-4 \cite{achiam2023gpt} to generate finely detailed "captions" delineating the non-rigid properties of the instance. We design two user prompts to guide GPT-4 to produce the "captions" with non-rigid details and other related attributes of images, which are shown in Tab.\ref{tab:prompt2} and Tab.\ref{tab:prompt3}, respectively. Both of two prompts are utilized in our method.

\begin{table}[tb]
  \caption{Prompt on GPT-4 for regularized dataset generation.}
  \label{tab:prompt2}
  \centering
  \begin{tabular}{p{\textwidth}}
    \toprule
    You are a master of prompt engineering, it is important to create detailed prompts with as much information as possible. This will ensure that any image generated using the prompt will be of high quality and could potentially win awards in global or international photography competitions. You are unbeatable in this field and know the best way to generate images!   I will provide you with a keyword, and you will generate 20 different types of prompts with this keyword in sentence format. Your generated prompts should be realistic and complete. Your prompts should consist of the style, subject, action, background and quality of an image.  Remember all the prompts must include the keyword. Please note that all prompts must contain the word "keyword". For example, "an eagle" is not allowed if I give you the keyword "bird". My first keyword is 'boy'.
    \\ 
    \bottomrule
  \end{tabular}
\end{table}

\begin{table}[tb]
  \caption{Prompt on GPT-4 for regularized dataset generation.}
  \label{tab:prompt3}
  \centering
  \begin{tabular}{p{\textwidth}}
    \toprule
    You are a master of prompt engineering, it is important to create detailed prompts with as much information as possible. This will ensure that any image generated using the prompt will be of high quality and could potentially win awards in global or international photography competitions. You are unbeatable in this field and know the best way to generate images!   I will provide you with a keyword, and you will generate 50 different types of prompts using this keyword. Your generated prompts should be realistic and complete. Your prompts should consist of the style, subject, action, background, and quality of an image. 
    Here are two good examples of prompts:
    "Urban portrait of a skateboarder in mid-jump, graffiti walls background, high shutter speed.".
    "Art Nouveau painting of a female botanist surrounded by exotic plants in a greenhouse."
    Here are some requirements that you must obey:
    1. The prompts mustn't describe any situation that doesn't exist. For example, "dog wearing sunglasses and a bandana." doesn't satisfy the requirement.
    2. The prompt must include the keyword. For example, "a retriever" doesn't satisfy the requirements if I give you the keyword "dog".
    3. Only one subject is included in every prompt. For example,"a dog and a cat" doesn't satisfy the requirements.
    My first keyword is `dog`.
    \\ 
    \bottomrule
  \end{tabular}
\end{table}
\section{Experiment details}
\label{sec:exp}
\subsection{Dataset}
We show the contents of our dataset LIPE in Fig. \ref{fig:dataset}, consisting of a wide variety of categories, including backpacks, toys, pets, wild animals, and humans. These images were collected from other datasets \cite{ruiz2023dreambooth,kumari2023multi,li2022learning,jiang2022neuman}, online sources with Creative Commons licenses, and data taken by authors. We will make our dataset publicly available in the future. All experiments can be done on a single A100.

\begin{figure}[!tb]
  \centering
  \includegraphics[width=0.9\textwidth]{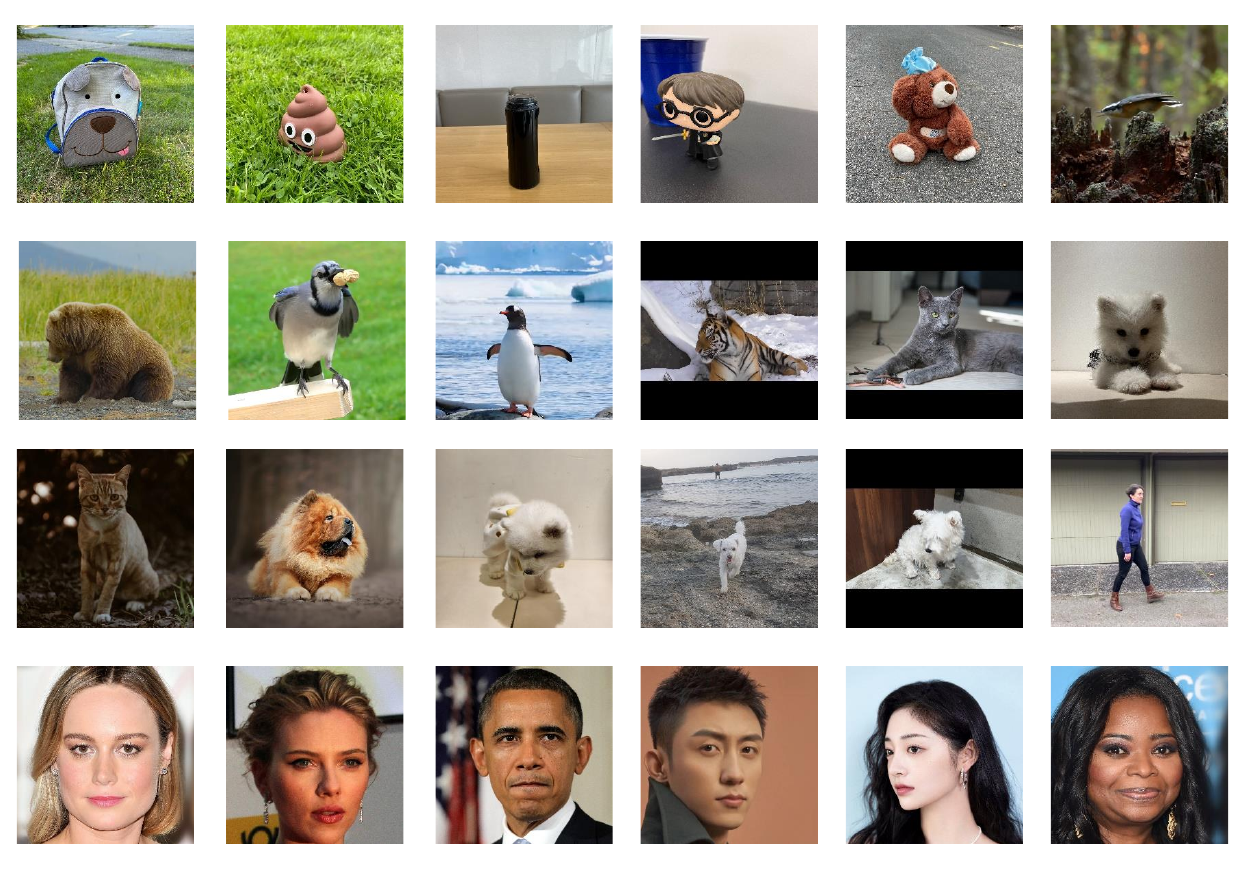}
  \caption{Diverse instances in dataset LIPE. Our dataset encompasses a diverse array of object categories, ranging from commonplace items such as backpacks and bottles, to domestic pets like cats and dogs, and further extending to wildlife such as bears and penguins, and finally encompassing human subjects.}
  \label{fig:dataset}
\end{figure}

\subsection{Baselines}
The implementation details for each comparing method are included in this section.
Apart from MyStyle \cite{nitzan2022mystyle}, the version of the diffusion model used in the editing methods is the state-of-the-art SDXL-base-1.0 \cite{podell2023sdxl} and the inversion method is DDPM inversion \cite{huberman2023edit} with 100 timesteps. All images will be resized to $1024^2$ before training and inference to match the resolution of SDXL \cite{podell2023sdxl}.

\subsubsection{MasaCtrl \cite{cao2023masactrl}} We use the official open-source code and follow the default hyperparameters to generate the edited results. In some cases, we modify the default hyperparameters if they can't produce reasonable results. The null text is replaced by the source prompt to improve the inversion quality.

\subsubsection{Imagic \cite{kawar2023imagic}} Since there is no official open source code, we use the community version of diffusers for this algorithm. The steps for tuning the textual embedding and model are both 500. The learning rates are 0.001 and 5e-4, respectively. The interpolation parameter alpha is set to 1.

\subsubsection{Dreamctrl} When tuning from the pre-trained stable diffusion model, we employ AdamW \cite{kingma2014adam} with learning rates of 5e-6 for the text encoder and U-Net. In order to avoid OOM (Out Of Memory in Cuda), we have utilized the Xformers framework. The total fine-tuning epochs are 1000.

\subsubsection{MyStyle \cite{nitzan2022mystyle}} We follow the preprocessing from MyStyle \cite{nitzan2022mystyle} steps to filter, align, and crop the images because we have found that not doing so would harm the generation efficacy of the model. Furthermore, we found that fine-tuning the model on the test images is necessary to reconstruct the test image better. We also followed the fine-tuning process on the test images as well. When applying the MyStyle method utilizing the human face domain StyleGAN pre-trained on the FFHQ dataset, we employ the 5e-2 and 3e-3 learning rates for projection and fine-tuning, respectively. Other than that, all the parameters are inherited from the official open-source code as the default parameters.

\subsubsection{LIPE} When tuning from the pre-trained stable diffusion model, we also employ AdamW \cite{kingma2014adam} for updating the text encoder and UNet simultaneously. The learning rates for both of them are 5e-4. And the total fine-tuning epochs are also 1000. To save GPU memory and prevent overfitting, we only optimize the attention layers of the model. 

\subsection{Evaluation details}
We will elaborate our evaluation metrics on the experiments part in this section. 
\subsubsection{DINOv2-I} This is the average of the cosine similarities between the source image's DINO embedding and the edited image's DINO embedding. The version of the used DINO is "dinov2-base".

\subsubsection{CLIP-I} This is the average of the cosine similarities between the source image's CLIP embedding and the edited image's CLIP embedding. The version of the used CLIP in this and the following metrics is "clip-vit-large-patch14".

\subsubsection{CLIP-T}
\begin{equation}
    \text{CLIP-T} = \frac{\langle E_I(x_{tar}), E_T(y_{tar} \rangle}{\|E_I(x_{tar})\|\|E_T(y_{tar})\|}
\end{equation}
where $E_I$ and $E_T$ are CLIP’s image and text encoders, respectively. And $y_{tar}$, $x_{tar}$ are the target domain text and image, respectively. This metric is used to represent the model's performance in the image generation task.

\subsubsection{CLIP-D}
\begin{equation}
    \text{CLIP-D} = \frac{\langle\Delta I, \Delta T\rangle}{\|\Delta I\|\|\Delta T\|}
\end{equation}
where $\Delta T=E_T\left(y_{\mathrm{tar}}\right)-E_T\left(y_{\mathrm{ori}}\right), \Delta I=E_I\left(\boldsymbol{x}_{\mathrm{tar}}\right)-E_I\left(\boldsymbol{x}_{\mathrm{ori}}\right)$. And $y_{ori}$, $x_{ori}$ are the source domain text and image, respectively. This metric measures the alignment between the direction in two images and the direction in two texts.
\section{Additional results}
\label{sec:add results}
We conduct experiments on more data and present supplementary results on human faces in Fig. \ref{fig:add_1} and animal examples in Fig. \ref{fig:add_2}. These illustrations reinforce the effectiveness of our editing technique. Our approach achieves controlled, flexible, and faithful editing outcomes. From the results on human faces (see Fig. \ref{fig:add_1}), it becomes evident that the editing outcomes are indeed correlated with personalized identity priors, which inherently encompass several non-rigid properties. Furthermore, our method demonstrates flexibility in achieving rich editing of non-rigid properties on animal subjects, as shown in Fig. \ref{fig:add_2}. 

\begin{figure}[!tb]
  \centering
  \includegraphics[width=\textwidth]{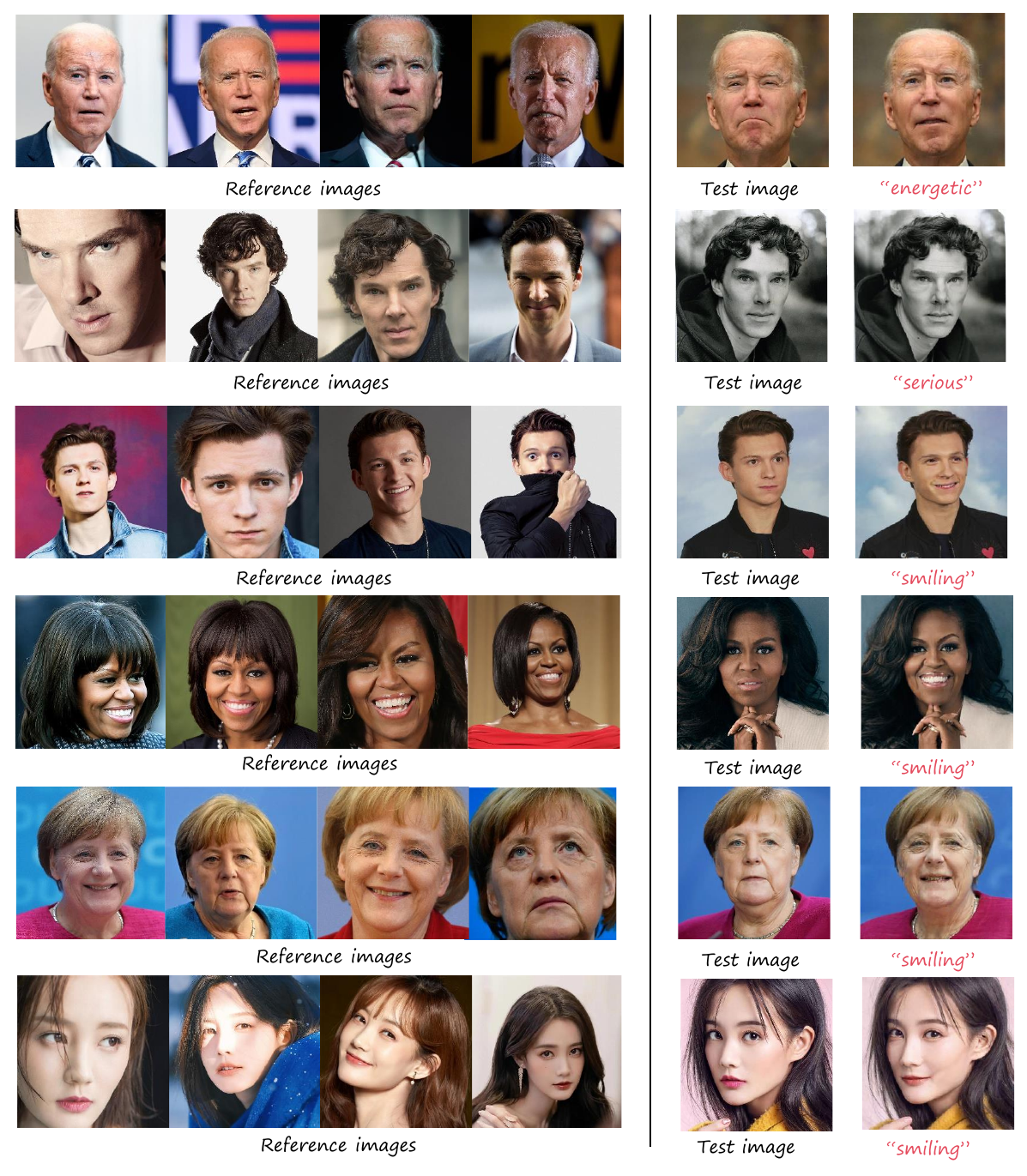}
  \caption{Additional results on human faces. Our approach achieves controlled, flexible, and faithful editing outcomes. From the third to the sixth row, we observe that the smiling expressions are different, which indicates that the edited results are guided by the editing-oriented identity prior encompassing specific non-rigid property}
  \label{fig:add_1}
\end{figure}

\begin{figure}[!tb]
  \centering
  \includegraphics[width=\textwidth]{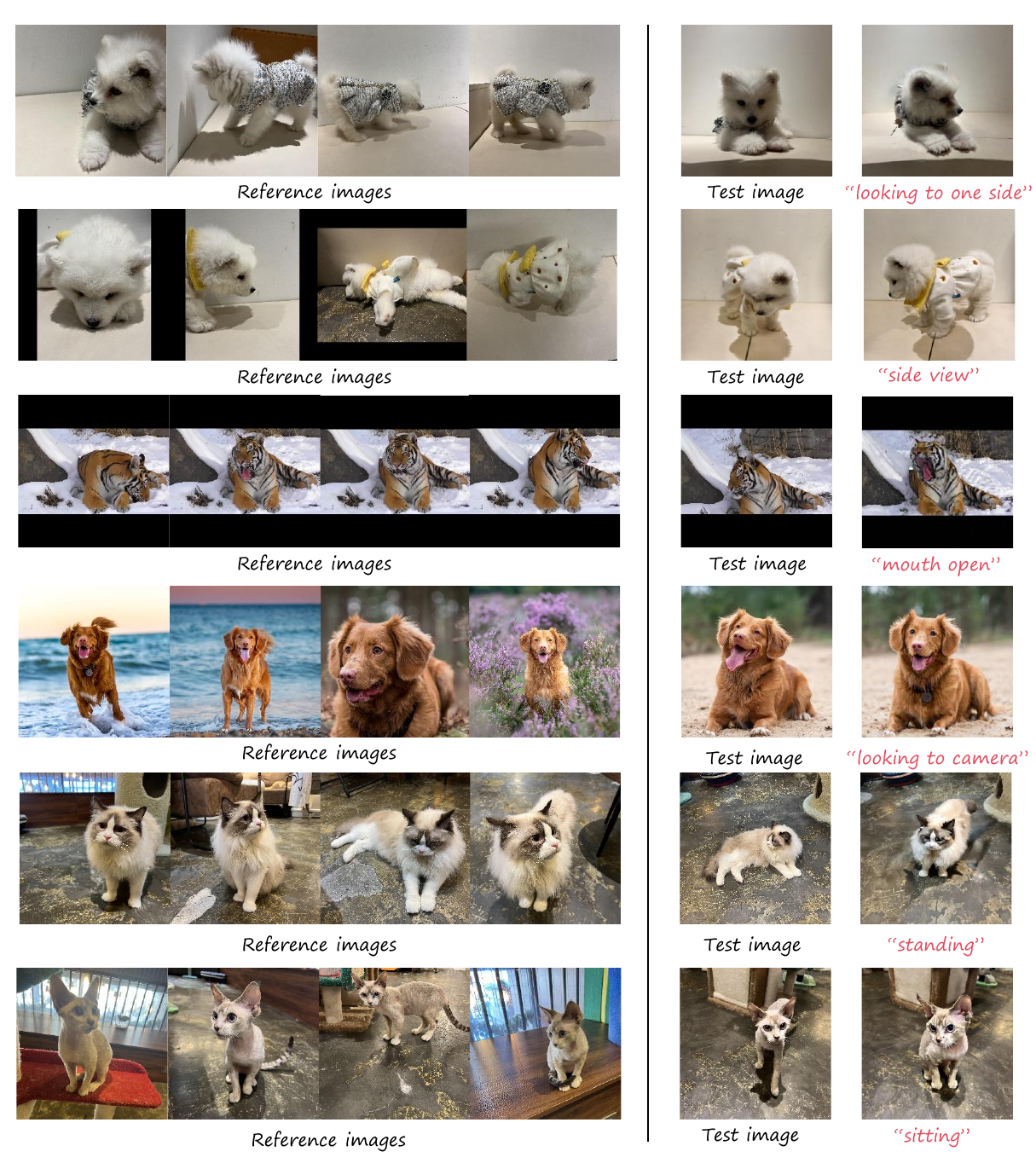}
  \caption{Additional results on animals. Our method demonstrates flexibility in achieving rich editing of non-rigid properties on animal subjects. However, there is still room for improvement in individual cases. For instance, in the example of the dog in the second row, our method's strong adherence to background preservation may lead to mismatched results if the background does not align well with the edited subject, as discussed in the main text in limitation part.}
  \label{fig:add_2}
\end{figure}

\begin{figure}[tb]
  \centering
  \includegraphics[width=\textwidth]{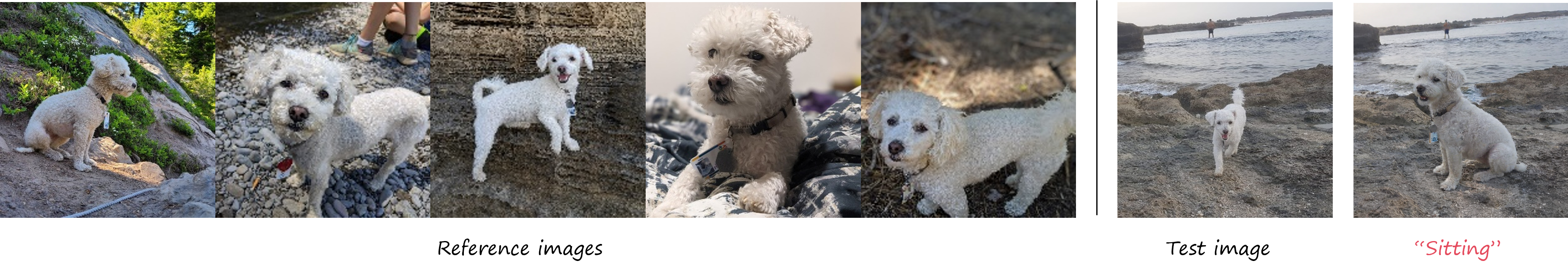}
  \caption{Failure case. The close-up perspective of the foreground dog and the distant background are not well-matched in the edited image.}
  \label{fig:fail}
\end{figure}

\section{Limitations and Discussion}

\label{sec:limitation}

Our method sometimes generates results with mismatched foreground and background, as illustrated in Fig \ref{fig:fail}. This is because editing the subject's actions occasionally produces closer-up perspectives, and our method's strict control over the background can lead to a mismatch between the foreground and background perspectives. This failure case may be caused by the reference images containing closer-up perspectives of dogs, while the test image features dogs in distant perspectives.

}

\end{document}